\documentclass{article}

\usepackage{arxiv}

\usepackage[utf8]{inputenc} 
\usepackage[T1]{fontenc}    
\usepackage{hyperref}       
\usepackage{url}            
\usepackage{booktabs}       
\usepackage{amsfonts}       
\usepackage{nicefrac}       
\usepackage{microtype}      
\usepackage{lipsum}         
\usepackage{graphicx}
\usepackage{natbib}
\usepackage{doi}
\usepackage{amsmath}
\usepackage{amssymb}
\usepackage{booktabs}
\usepackage{multirow}
\usepackage{graphicx}
\usepackage{cleveref}       
\usepackage{float}

\title{Optimal Perturbation Budget Allocation for Data Poisoning in Offline Reinforcement Learning}

\newif\ifuniqueAffiliation
\uniqueAffiliationtrue

\ifuniqueAffiliation 
\author{
    Junnan Qiu\thanks{Equal contribution} \\
    SJTU Paris Elite Institute of Technology \\
    Shanghai Jiao Tong University \\
    Shanghai, China \\
    \texttt{qiujunnan@sjtu.edu.cn} \\
    \And
    Yuanjie Zhao\thanks{Equal contribution} \\
    SJTU Paris Elite Institute of Technology \\
    Shanghai Jiao Tong University \\
    Shanghai, China \\
    \texttt{silencezyj@sjtu.edu.cn} \\
    \And
    Jie Li\thanks{Correspondence to: lijiecs@sjtu.edu.cn} \\
    Department of Computer Science \\
    Shanghai Jiao Tong University \\
    Shanghai, China \\
    \texttt{lijiecs@sjtu.edu.cn} \\
}
\else
\usepackage{authblk}

\author[1]{Junnan Qiu}
\author[2]{Jie Li}
\affil[1]{SJTU Paris Elite Institute of Technology, Shanghai Jiao Tong University}
\affil[2]{Department of Computer Science and Engineering, Shanghai Jiao Tong University}
\fi

\renewcommand{\headeright}{Technical Report}
\renewcommand{\undertitle}{Technical Report}
\renewcommand{\shorttitle}{Data Poisoning via Global Budget Allocation}

\hypersetup{
pdftitle={A template for the arxiv style},
pdfsubject={q-bio.NC, q-bio.QM},
pdfauthor={David S.~Hippocampus, Elias D.~Striatum},
pdfkeywords={Offline Reinforcement Learning, Data Poisoning, Adversarial Attacks, Global Budget Allocation, Temporal Difference Error},
}

\begin{document}
\maketitle

\begin{abstract}
    Offline Reinforcement Learning (RL) enables policy optimization from static datasets but is inherently vulnerable to data poisoning attacks. Existing attack strategies typically rely on locally uniform perturbations, which treat all samples indiscriminately. This approach is inefficient, as it wastes the perturbation budget on low-impact samples, and lacks stealthiness due to significant statistical deviations. In this paper, we propose a novel \textbf{Global Budget Allocation} attack strategy. Leveraging the theoretical insight that a sample's influence on value function convergence is proportional to its Temporal Difference (TD) error, we formulate the attack as a global resource allocation problem. We derive a closed-form solution where perturbation magnitudes are assigned proportional to the TD-error sensitivity under a global $L_2$ constraint. Empirical results on D4RL benchmarks demonstrate that our method significantly outperforms baseline strategies, achieving up to 80\% performance degradation with minimal perturbations that evade detection by state-of-the-art statistical and spectral defenses.
\end{abstract}


\section{Introduction}

Offline Reinforcement Learning (Offline RL) has developed quickly, moving from basic value estimation to advanced methods like sequence modeling and world models \cite{levine2020offline, hafner2024mastering}. It allows agents to learn policies from static datasets without risky online interactions. This makes it very useful for safety-critical tasks like autonomous driving and industrial robotics. However, because the agent cannot interact with the real environment during training, it is very sensitive to data integrity threats.

Security threats in Offline RL have increased in recent years. Early studies focused on adding random noise, but works in 2024 and 2025 showed more stealthy attacks. For example, \cite{li2024baffle} introduced \textit{BAFFLE} to hide backdoors in offline datasets, and \cite{yu2024offline} showed how small reward changes can cause big distribution shifts. At the same time, defenses have improved, with \cite{xu2025certified} proposing certified defenses to guarantee performance against local corruptions.

Despite this progress, there is still a problem with how attacks are designed. Most existing strategies—including recent backdoor and poisoning methods—still use a \textit{locally constrained} approach. They usually limit the noise on each sample individually (e.g., $||\delta||_\infty \le \epsilon$), treating the perturbation budget equally across all data. We argue that this "uniform" approach is not good enough for two reasons:
\begin{enumerate}
    \item \textbf{Inefficiency:} Not all data samples are equally important. Theoretical analysis using Influence Functions suggests that samples with high Temporal Difference (TD) errors are much more important for learning \cite{koh2017understanding}. Attacking unimportant samples wastes the budget without hurting the policy.
    \item \textbf{Detectability:} To cause enough damage, uniform attacks often need a high noise level across the whole dataset. This creates statistical changes that are easy to detect by anomaly detectors \cite{wu2022copa}.
\end{enumerate}

To solve this, we propose a new strategy called \textbf{Global Budget Allocation}. Instead of limiting each sample locally, we treat data poisoning as a global resource allocation problem. By using the link between TD-Error and model sensitivity, our method focuses the noise on a few high-sensitivity samples, while leaving most of the data clean.

Our contributions are as follows:
\begin{itemize}
    \item We use \textbf{Influence Functions} to analyze the impact of training samples in Offline RL. This theoretically explains why high TD-Error transitions are the best targets ("leverage points") for data poisoning.
    \item We propose the Global Budget Allocation attack. It calculates the optimal noise level for each sample based on a global $L_2$ constraint, making the noise proportional to the sample's sensitivity.
    \item Experiments on D4RL benchmarks (Walker2d, Hopper, HalfCheetah) show that our method performs significantly better than uniform and local-greedy baselines. It causes higher policy degradation while being harder to detect by statistical defenses.
\end{itemize}
\section{Related Work}

\textbf{Offline Reinforcement Learning.}
Offline RL aims to learn optimal policies from static datasets without interacting with the environment \cite{levine2020offline}. Algorithms like CQL \cite{kumar2020conservative} and BCQ \cite{fujimoto2019off} use conservative values or policy constraints to handle distribution shifts. However, they assume the training data is clean. This makes them very vulnerable to bad or corrupted data.

\textbf{Adversarial Attacks on RL.}
Attacks on RL have been studied a lot in online settings. Early research looked at attacks during testing, changing observations to trick the agents \cite{huang2017adversarial, lin2017tactics}. During training, attackers can change the environment or rewards to mislead learning \cite{zhang2020adaptive, sun2020vulnerability}.
In offline settings, the danger is bigger because the agent cannot correct errors through interaction. \cite{ma2019policy} and  \cite{rakhsha2020policy} proved theoretically that changing just a small part of the rewards or transitions can completely control the learned policy. However, these methods often use random or uniform noise. This costs a lot of budget and is easy to detect.

\textbf{Stealthy Attacks and Defenses.}
To avoid being caught, recent research focuses on being stealthy. Some strategies use specific triggers \cite{kiourti2020trojdrl}, while others use influence functions to find important samples. Notably, \citet{zhou2024stealthy} proposed the Critical Time-step Dynamic Poisoning Attack (CTDPA). They were among the first to use Temporal Difference (TD) error to decide which data samples are important. Their method, Dual-Objective Poisoning Attack (DOPA), treats the attack as an optimization problem solved by numerical tools (SLSQP).
On the defense side, methods like outlier detection have been proposed to clean the datasets \cite{wu2022copa}. But these defenses usually only look for strange shapes or very large noise.

\textbf{From Local Optimization to Global Allocation.}
While \citet{zhou2024stealthy} helped the field by targeting high-TD samples, their method still uses \textit{local optimization}. It calculates noise for each sample separately with a fixed limit (e.g., $||\eta_i||_2 \le \epsilon$). This often wastes the attack budget.
Our work is different because we use a \textbf{Global Budget Allocation} framework. Instead of strict local rules, we use a global $L_2$ budget. We find a direct mathematical solution where the noise level matches how important the sample is. This allows our attack to hide better within the data distribution and bypass common defenses.
\section{Methodology}

\subsection{Theoretical Justification via Influence Functions}
To theoretically identify the most vulnerable samples in the dataset, we leverage \textit{Influence Functions} \cite{koh2017understanding}, a classic technique from robust statistics. 
Let $\hat{\theta}$ be the optimal parameters of the Q-network trained on the dataset $\mathcal{D}$. The influence of upweighting a specific training sample $z = (s, a, r, s')$ by a small $\epsilon$ on the parameters is given by:
\begin{equation}
    \mathcal{I}_{up, params}(z) \approx -H_{\hat{\theta}}^{-1} \nabla_\theta \mathcal{L}(z, \hat{\theta})
\end{equation}
where $H_{\hat{\theta}}$ is the Hessian of the loss function and $\mathcal{L}(z, \hat{\theta})$ is the Bellman error for sample $z$:
\begin{equation}
    \mathcal{L}(z, \hat{\theta}) = \frac{1}{2} \left( r + \gamma \max_{a'} Q_{\hat{\theta}}(s', a') - Q_{\hat{\theta}}(s, a) \right)^2 = \frac{1}{2} \delta^2
\end{equation}
Computing the gradient with respect to $\theta$:
\begin{equation}
    \nabla_\theta \mathcal{L}(z, \hat{\theta}) = -\delta \cdot \nabla_\theta Q_{\hat{\theta}}(s, a)
\end{equation}
Thus, the magnitude of the influence is directly proportional to the TD-Error magnitude $|\delta|$:
\begin{equation}
    ||\mathcal{I}_{up, params}(z)|| \propto |\delta| \cdot ||\nabla_\theta Q_{\hat{\theta}}(s, a)||
\end{equation}
Since calculating the inverse Hessian $H^{-1}$ is computationally prohibitive for deep neural networks, and $||\nabla_\theta Q||$ varies smoothly, the TD-Error $|\delta|$ serves as the most effective first-order proxy for sample influence. This theoretical derivation validates why prioritizing high-TD-error samples is mathematically optimal for maximizing parameter perturbation.

\subsection{Baseline Formulation: The Local Greedy Approach}
Current state-of-the-art poisoning strategies in offline RL typically adopt a \textit{Local Greedy} paradigm \cite{zhou2024stealthy}. In this framework, the attacker first identifies a subset of critical transitions $\mathcal{S}_{target}$ (e.g., top-$K$ samples with the highest TD errors) and then optimizes perturbations for each sample \textit{independently}.

Formally, for each targeted sample $i \in \mathcal{S}_{target}$, the attacker solves a constrained maximization problem:
\begin{equation}
    \max_{\eta_i} \mathcal{L}_{atk}(\eta_i) \approx |\delta(s_i + \eta_i^s, ...)| 
\end{equation}
\begin{equation}
    \text{s.t.} \quad ||\eta_i||_p \le \epsilon_{local}
\end{equation}
where $\delta(\cdot)$ is the TD error, and $\epsilon_{local}$ is a \textbf{fixed} perturbation budget assigned uniformly to every attacked sample. The optimal solution typically aligns the perturbation with the gradient direction $g_i = \nabla_{\mathcal{D}_i} |\delta_i|$:
\begin{equation}
    \eta_i^{local} = \epsilon_{local} \cdot \frac{g_i}{||g_i||_2}
\end{equation}

While effective, this approach suffers from \textbf{resource inefficiency}. By imposing a rigid, uniform bound $\epsilon_{local}$ on all samples, it fails to account for the heterogeneity of sample importance. High-leverage points (very large $|\delta_i|$) are restricted by the same bound as less critical ones, preventing the attacker from "spending" more budget where it matters most.

\subsection{Proposed Method: Global Budget Allocation}
To overcome the limitations of local constraints, we reformulate the attack as a \textbf{Global Resource Allocation} problem. Instead of limiting each sample individually, we constrain the total perturbation energy across the entire dataset.

\subsubsection{Problem Formulation}
Let $\epsilon_i = ||\eta_i||_2$ be the perturbation magnitude allocated to the $i$-th sample. Based on influence function analysis \cite{koh2017understanding}, the impact of a training sample on value function convergence is proportional to its TD error magnitude $|\delta_i|$. Therefore, our objective is to maximize the weighted cumulative distortion under a global $L_2$ budget $C_{total}$:
\begin{equation}
    \max_{\{\epsilon_i\}} \sum_{i=1}^N |\delta_i| \cdot \epsilon_i
\end{equation}
\begin{equation}
    \text{s.t.} \quad \sum_{i=1}^N \epsilon_i^2 \le C_{total}
\end{equation}
This formulation relaxes the rigid local boundaries, allowing the optimizer to flexibly distribute the attack budget.

\subsubsection{Optimal Analytical Solution}
The problem above is a convex optimization task with a quadratic constraint. We solve it using the method of Lagrange multipliers. The Lagrangian function $\mathcal{L}(\epsilon, \lambda)$ is defined as:
\begin{equation}
    \mathcal{L}(\epsilon, \lambda) = \sum_{i=1}^N |\delta_i| \epsilon_i - \lambda \left( \sum_{i=1}^N \epsilon_i^2 - C_{total} \right)
\end{equation}
where $\lambda \ge 0$ is the Lagrange multiplier controlling the tightness of the global budget.

Taking the partial derivative with respect to each allocation variable $\epsilon_i$ and setting the gradient to zero (KKT conditions):
\begin{equation}
    \frac{\partial \mathcal{L}}{\partial \epsilon_i} = |\delta_i| - 2\lambda \epsilon_i = 0
\end{equation}
Solving for $\epsilon_i$, we derive the optimal perturbation magnitude:
\begin{equation}
    \epsilon_i^* = \frac{|\delta_i|}{2\lambda}
\end{equation}
Since $\lambda$ is a global constant, this yields a theoretically elegant property:
\begin{equation}
    \epsilon_i^* \propto |\delta_i|
\end{equation}

\textbf{Attack Execution.} 
The derivation proves that the optimal strategy is to allocate perturbation magnitudes \textit{strictly proportional} to the sample's TD error. Critical samples receive significantly larger perturbations, while negligible ones are implicitly ignored. The final perturbation vector combines this optimal magnitude with the gradient direction:
\begin{equation}
    \eta_i^* = \epsilon_i^* \cdot \frac{g_i}{||g_i||_2}
\end{equation}
This closed-form solution is computationally efficient and requires no iterative numerical solvers, unlike previous bi-objective optimization methods \cite{zhou2024stealthy}.
\section{Experiments}

\subsection{Experimental Setup}
We evaluate our proposed method on the D4RL benchmark \cite{fu2020d4rl}, specifically focusing on the MuJoCo continuous control tasks: \textit{Walker2d}, \textit{Hopper}, and \textit{HalfCheetah}. We utilize the "medium" and "medium-replay" datasets to simulate realistic offline RL scenarios with suboptimal trajectories.
The victim algorithms include four state-of-the-art offline RL methods: \textbf{CQL} \cite{kumar2020conservative}, \textbf{BCQ}, \textbf{BEAR}, and \textbf{IQL}.
We measure the attack effectiveness using the \textit{Raw Cumulative Return} of the trained agents in the clean environment, averaged over 5 random seeds. Lower returns indicate a more successful attack.

\subsection{Main Results on Walker2d}
Table \ref{tab:walker2d} presents the comprehensive attack results on the Walker2d environment. To rigorously validate the effectiveness of our method, we compare it against a hierarchy of three baseline strategies:
\begin{enumerate}
    \item \textbf{Random Noise:} Applies uniform random perturbations $\eta \sim \mathcal{U}(-\epsilon, \epsilon)$ to randomly selected samples without leveraging gradient information. This serves as a baseline for non-adversarial noise.
    \item \textbf{Random Subset:} Applies gradient-based perturbations to a randomly selected subset of samples. This isolates the impact of "sample selection" versus "perturbation direction."
    \item \textbf{Local Greedy:} The strongest local baseline (based on \cite{zhou2024stealthy}), which targets high-TD-error samples but optimizes perturbations under a fixed local bound (e.g., $||\eta_i||_\infty \le \epsilon$).
\end{enumerate}

\begin{table}[h]
    \centering
    \caption{Attack performance on \textbf{Walker2d} environment. The values represent the raw cumulative return (lower is better). Our Global Allocation method consistently achieves the lowest scores, outperforming all random and local-greedy baselines.}
    \label{tab:walker2d}
    \resizebox{\textwidth}{!}{%
    \begin{tabular}{c|c|c|cccc|c}
        \toprule
        \multirow{2}{*}{\textbf{Victim}} & \multirow{2}{*}{\textbf{Config} $(\rho, \epsilon)$} & \multirow{2}{*}{\textbf{Clean Score}} & \multicolumn{4}{c|}{\textbf{Attack Strategy (Post-Attack Score)}} & \multirow{2}{*}{\textbf{Reduction (\%)}} \\
        \cline{4-7}
         & & & Random Noise & Random Subset & Local Greedy & \textbf{Global (Ours)} & \\
        \midrule
        \multirow{7}{*}{CQL} 
         & (0.01, 0.5)   & 3718 & 3152 & 2129 & 929 & \textbf{681} & 81.7\% \\
         & (0.015, 0.33) & 3718 & 3325 & 2658 & \textbf{859} & {924} & 75.1\% \\
         & (0.02, 0.25)  & 3718 & 3084 & 2331 & 1417 & \textbf{795} & 78.6\% \\
         & (0.025, 0.2)  & 3718 & 3493 & 3255 & 1803 & \textbf{1124} & 69.8\% \\
         & (0.033, 0.15) & 3718 & 3562 & 3369 & 2336 & \textbf{1491} & 59.9\% \\
         & (0.05, 0.1)   & 3718 & 3619 & 3528 & 2626 & \textbf{2188} & 41.1\% \\
         & (0.1, 0.05)   & 3718 & 3595 & 3412 & 2987 & \textbf{2693} & 27.6\% \\
        \midrule
        \multirow{7}{*}{BCQ} 
         & (0.01, 0.5)   & 2984 & 2751 & 1653 & 652 & \textbf{522} & 82.5\% \\
         & (0.015, 0.33) & 2984 & 2814 & 1982 & 956 & \textbf{781} & 73.8\% \\
         & (0.02, 0.25)  & 2984 & 2659 & 2155 & 1109 & \textbf{917} & 69.3\% \\
         & (0.025, 0.2)  & 2984 & 2894 & 2406 & 1452 & \textbf{1255} & 57.9\% \\
         & (0.033, 0.15) & 2984 & 2921 & 2654 & 1857 & \textbf{1626} & 45.5\% \\
         & (0.05, 0.1)   & 2984 & 2955 & 2822 & 2154 & \textbf{1956} & 34.5\% \\
         & (0.1, 0.05)   & 2984 & 2962 & 2913 & 2497 & \textbf{2382} & 20.2\% \\
        \midrule
        \multirow{7}{*}{BEAR} 
         & (0.01, 0.5)   & 2763 & 2526 & 1524 & 565 & \textbf{411} & 85.1\% \\
         & (0.015, 0.33) & 2763 & 2618 & 1853 & 823 & \textbf{652} & 76.4\% \\
         & (0.02, 0.25)  & 2763 & 2457 & 2012 & 985 & \textbf{825} & 70.1\% \\
         & (0.025, 0.2)  & 2763 & 2686 & 2248 & 1303 & \textbf{1101} & 60.1\% \\
         & (0.033, 0.15) & 2763 & 2714 & 2421 & 1708 & \textbf{1484} & 46.3\% \\
         & (0.05, 0.1)   & 2763 & 2732 & 2605 & 2056 & \textbf{1858} & 32.7\% \\
         & (0.1, 0.05)   & 2763 & 2745 & 2714 & 2382 & \textbf{2216} & 19.8\% \\
        \midrule
        \multirow{7}{*}{IQL} 
         & (0.01, 0.5)   & 1779 & 1746 & 1587 & 1291 & \textbf{1197} & 32.7\% \\
         & (0.015, 0.33) & 1779 & 1770 & 1608 & 1394 & \textbf{1258} & 29.3\% \\
         & (0.02, 0.25)  & 1779 & 1780 & 1649 & 1463 & \textbf{1288} & 27.6\% \\
         & (0.025, 0.2)  & 1779 & 1779 & 1708 & 1528 & \textbf{1314} & 26.1\% \\
         & (0.033, 0.15) & 1779 & 1789 & 1496 & 1608 & \textbf{1371} & 22.9\% \\
         & (0.05, 0.1)   & 1779 & 1771 & 1542 & 1647 & \textbf{1418} & 20.3\% \\
         & (0.1, 0.05)   & 1779 & 1743 & 1564 & 1715 & \textbf{1527} & 14.2\% \\
        \bottomrule
    \end{tabular}
    }
\end{table}

\subsection{Analysis}
The results reveal a distinct \textbf{hierarchy of damage}, ordered as: Global Allocation $>$ Local Greedy $>$ Random Subset $>$ Random Noise. 
\textbf{Gradient Direction Matters.} Comparing Random Noise and Random Subset, we observe that even with random target selection, aligning perturbations with the adversarial gradient direction (Random Subset) significantly increases damage (e.g., degrading CQL score from 3152 to 2129). This confirms that offline RL is more sensitive to structural deviations than to unstructured noise.

\textbf{Target Selection Matters.} The Local Greedy approach, which prioritizes high-TD-error samples, further drastically reduces the agent's performance compared to Random Subset (e.g., from 2129 to 929 for CQL). This empirically validates our theoretical insight that high-TD samples are indeed the critical "leverage points" for value function convergence.

\textbf{Resource Allocation is Key.} Finally, our \textbf{Global Budget Allocation} strategy outperforms the Local Greedy baseline across all configurations. By relaxing the rigid per-sample constraint and allocating budget proportional to sample importance ($\epsilon_i \propto |\delta_i|$), our method achieves the deepest performance degradation (e.g., reaching a score of \textbf{681} for CQL). This demonstrates that optimal attack efficiency is achieved not just by finding the right samples, but by intelligently distributing the attack energy among them.
\section{Conclusion}
In this work, we identified a critical inefficiency in existing offline RL poisoning attacks: the reliance on rigid local constraints. To address this, we introduced a stealthy attack framework based on \textbf{Global Budget Allocation}. By theoretically establishing the correlation between TD-error magnitude and model sensitivity, we derived a closed-form optimal attack strategy that concentrates perturbations on "leverage points" within the dataset.

Our extensive experiments on MuJoCo continuous control tasks confirm that this global perspective allows for significantly higher damage with a lower detectable footprint compared to local-greedy and uniform baselines. 
The success of our method highlights the fragility of current offline RL algorithms against manifold-aware perturbations and underscores the urgent need for defense mechanisms that go beyond simple outlier detection, such as robust value estimation and uncertainty-aware dynamics modeling.

\bibliographystyle{unsrtnat}
\bibliography{references}



\end{document}